\documentclass[a4paper,conference]{IEEEtran}

\usepackage{ifpdf}
\usepackage{cite}
\usepackage[pdftex]{graphicx}
\usepackage{array}
\usepackage{mdwmath}
\usepackage{mdwtab}
\usepackage{amssymb,latexsym}
\usepackage{stfloats}
\usepackage{amsmath}
\usepackage{cite}
\usepackage{color}
\usepackage{algcompatible}
\usepackage{caption}
\usepackage{graphicx}
\usepackage{algorithm}
\usepackage{subcaption}
\usepackage[font={footnotesize}]{caption}

\usepackage{balance}

\usepackage[utf8]{inputenc}

\usepackage[T1]{fontenc}

\usepackage[compatible]{algpseudocode}

\renewcommand{\figurename}{Figure}

\DeclareRobustCommand*{\IEEEauthorrefmark}[1]{%
\raisebox{0pt}[0pt][0pt]{\textsuperscript{\footnotesize\ensuremath{#1}}}}

\renewcommand\IEEEkeywordsname{Keywords}

\makeatletter
\def\ps@IEEEtitlepagestyle{%
  \def\@oddfoot{\mycopyrightnotice}%
  \def\@evenfoot{}%
}
\def\mycopyrightnotice{%
  \footnotesize
  \vspace*{40pt}
  \parbox{\textwidth}{%
    \centering
    \fbox{%
      \begin{minipage}{0.9\textwidth}
        © 2023 IEEE.  Personal use of this material is permitted.  Permission from IEEE must be obtained for all other uses, in any current or future media, including reprinting/republishing this material for advertising or promotional purposes, creating new collective works, for resale or redistribution to servers or lists, or reuse of any copyrighted component of this work in other works.
      \end{minipage}%
    }%
  }%
  \gdef\mycopyrightnotice{}
}
\begin{document}
\title{SEMI-DiffusionInst: A Diffusion Model Based Approach for Semiconductor Defect Classification and Segmentation} 
\author{\IEEEauthorblockN{
Vic De Ridder\IEEEauthorrefmark{1}$^,$\IEEEauthorrefmark{2}*,
Bappaditya Dey\IEEEauthorrefmark{2}*,
Sandip Halder\IEEEauthorrefmark{2}} \& Bartel Van Waeyenberge\IEEEauthorrefmark{1}
\IEEEauthorblockA{\IEEEauthorrefmark{1}
Ghent University,9000 Ghent, Belgium}
\IEEEauthorblockA{\IEEEauthorrefmark{2}
Interuniversity Microelectronics Centre (imec), 3001 Leuven, Belgium}
*Vic De Ridder and Bappaditya Dey contributed equally to the work. \\
{\it Bappaditya.Dey@imec.be}}

\maketitle

\begin{abstract}
With continuous progression of Moore's Law, integrated circuit (IC) device complexity is also increasing. Scanning Electron Microscope (SEM) image based extensive defect inspection and accurate metrology extraction are two main challenges in advanced node (2 nm and beyond) technology. Deep learning (DL) algorithm based computer vision approaches gained popularity in semiconductor defect inspection over last few years. In this research work, a new semiconductor defect inspection framework "SEMI-DiffusionInst" is investigated and compared to previous frameworks. To the best of the authors' knowledge, this work is the first demonstration to accurately detect and precisely segment semiconductor defect patterns by using a diffusion model. Different feature extractor networks as backbones and data sampling strategies are investigated towards achieving a balanced trade-off between precision and computing efficiency. Our proposed approach outperforms previous work on overall mAP and performs comparatively better or as per for almost all defect classes (per class APs). The bounding box and segmentation mAPs achieved by the proposed SEMI-DiffusionInst model are improved by 3.83\% and 2.10\%,respectively. Among individual defect types, precision on line collapse and thin bridge defects are improved approximately 15\% on detection task for both defect types. It has also been shown that by tuning inference hyperparameters, inference time can be improved significantly without compromising model precision. Finally, certain limitations and future work strategy to overcome them are discussed.
\end{abstract}

\begin{IEEEkeywords}
Metrology, defect inspection, scanning electron microscope, defect area segmentation, defect classification, deep learning, DDPM
\end{IEEEkeywords}

\IEEEpeerreviewmaketitle

\section{Introduction}
\begin{figure}[b!]
\center
\includegraphics[width=0.15\textwidth]{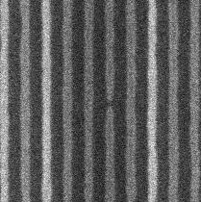}
\hspace{0.3em}
\includegraphics[width=0.15\textwidth]{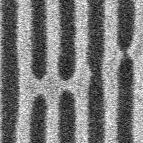}
\hspace{0.3em}
\includegraphics[width=0.15\textwidth]{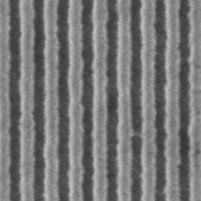}
\caption{Examples of challenging semiconductor defects. From left to right: Nano-gap, multi-bridges and micro-bridges.}
\label{subfigure_example}
\end{figure}

Production of a chip from a raw silicon wafer passes through different process steps such as doping, litho, etching, chemical mechanical polishing and electrical testing. Progression of Moore's law  in the realm of advanced node technology (2nm and beyond), has challenged conventional metrology and process precision. High-NA (numerical aperture) EUV (Extreme ultraviolet lithography), emerged as a possible solution to print smaller features, require thinner resist. 
This leads to multifaceted challenging problems such as (a) introduction of dynamic stochastic defects at nano-scale range, (b) image contrast loss and (c) depreciated SNR (Signal-to-Noise ratio). Fig. \ref{subfigure_example} shows some of the challenging semiconductor defect types. Defect detection, classification and feature extraction are crucial steps in semiconductor processing. Pixel wise segmentation of a defect can be considered the most detailed visual classification technique as geometrical attributes (like polygon shape and size) can be completely extracted. These features can then be used to group different defect patterns at intra and inter class level. 
Conventional state-of-the-art defect detection tools (optical/e-beam) as well as statistical algorithms are generally failing at requirements of the semiconductor industry at these advanced nodes. Against this, Deep learning (DL) algorithm based computer vision approaches gained popularity in semiconductor defect inspection over last few years with continuous innovations in Convolutional Neural Network (CNN) architectures \cite{alexnet,yolo,swin}. This work investigates a new DL based framework for semiconductor (SEM based) defect detection and segmentation, which makes use of the diffusion generative AI (Artificial Intelligence) paradigm. The main contributions of this research work are: (i) a new semiconductor defect inspection framework "SEMI-DiffusionInst", based on \cite{diffinst}, is investigated and compared to previous frameworks. To the best of the authors’ knowledge, this work is the first demonstration to accurately detect and precisely segment semiconductor defect patterns by using a diffusion model. (ii) Benchmarking with different feature extractor networks as backbones with different data sampling strategies towards achieving a balanced trade-off between precision and computing efficiency. (iii) Investigating inference hyperparameter(s) tuning towards improving inference time without compromising model precision.

\begin{figure*}[t]

    \centering
    \begin{minipage}{\textwidth}
           \includegraphics[width=0.19\linewidth]{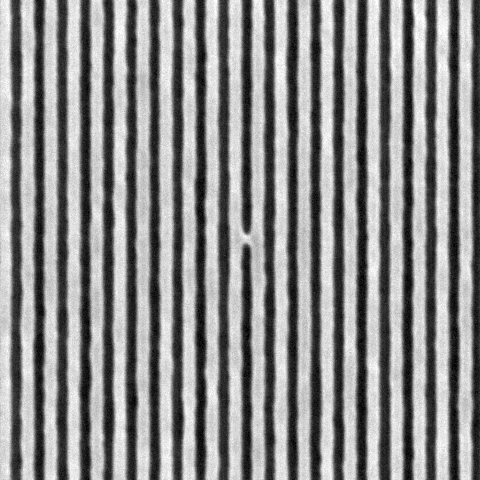} 
\hfill
    \includegraphics[width=0.19\linewidth]{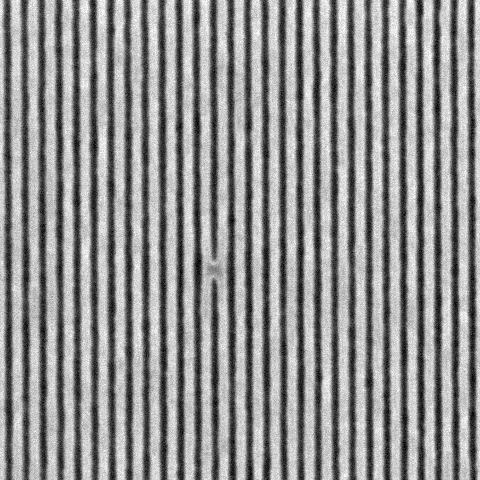}
    \hfill
        \includegraphics[width=0.19\linewidth]{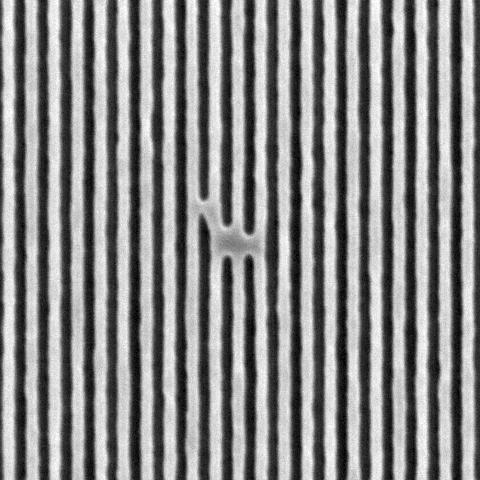}
\hfill
        \includegraphics[width=0.19\linewidth]{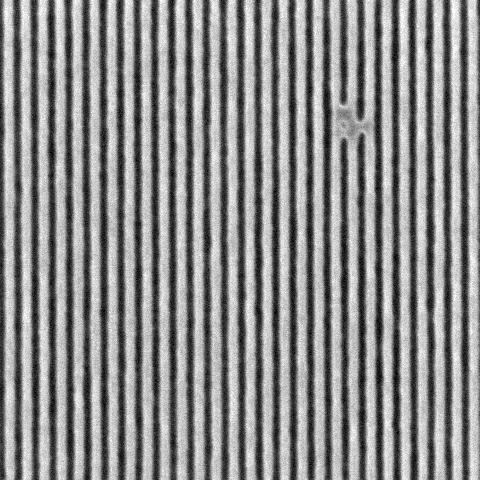}
\hfill
    \includegraphics[width=0.19\linewidth]{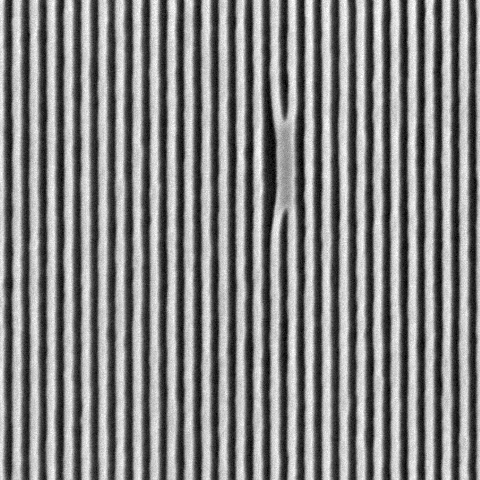}
        \caption{SEM image examples from used dataset for different defect types. From left to right: thin bridge, single bridge, multi bridge (non-horizontal), multi bridge (horizontal), line collapse.}
        \label{exampledataset}
    \end{minipage}
\end{figure*}
\section{Previous work}
\subsection{Semiconductor defect inspection}
In this section, selected previous research works in the context of semiconductor defect inspection will be discussed. A CNN based classification model \cite{cnnvsknn} is formulated to classify and detect wafer defects in SEM images by sliding the classifier over patches of the input image. The technique can also classify defects not seen during training by applying the k nearest neighbours algorithm to the classifier's feature space. This study demonstrated relatively better performance of a CNN approach towards defect inspection compared to other machine learning based approaches. However, the approach uses a predetermined sliding window, resulting in inflexibility towards variations in defect size.

Precise pixel-based multi-class defect pattern segmentation from inherently noisy SEM images towards generating a defect mask is a challenging task. This research work \cite{segrcnn} investigated the use of Mask R-CNN approach \cite{ormaskrcnn} towards precise defect pattern segmentation task. Dey et al. proposed SEMI-PointRend model \cite{pointrend} to overcome the limitations of previous work by replacing the segmentation network used on each proposed region with a PointRend \cite{orpointrend} module. PointRend predicts a segmentation mask for a given area by iteratively refining the segmentation and in this way it produces masks with more accurate edges, which is crucial for defect (polygon) patterns. Precise segmentation helped towards advanced statistical analysis of different defect classes as well as multiple instances of the same class.

In a very recent research work \cite{quantum}, authors demonstrated defect learning defect review (DLDR) by leveraging quantum computing based information processing. They proposed a classical-quantum hybrid DL algorithm on near-term quantum processors. Proposed framework driven quantum circuit enables wafer defect map classification, defect pattern classification, and hotspot detection by fine-tuning algorithm parameters.

\subsection{Denoising Diffusion Probabilistic models}
Denoising Diffusion Probabilistic Models (DDPM) were first introduced for synthetic image generation by \cite{ddpm}. However, the method can be applied to a variety of generative problems on the condition that the data can be represented in a continuous space. It treats the data generation problem as a gradual denoising process. First, a diffusive process is constructed which turns a clean data sample $x_0$ into standard normally distributed noise using a series of steps by sampling a random variable $n$ from the standard normal distribution at each step:
\begin{equation}
\label{classicdif}
    x_t = \sqrt{1-\beta_t}\cdot x_{t-1} + \sqrt{\beta_t} \cdot n
\end{equation}
In the original work \cite{ddpm}, the sequence $\beta_{1:T}$ was linearly increasing such that at the end of the diffusion process (the final timestep $T$), $x_T$ can be regarded as drawn from a standard normal distribution. Since a linear combination of normally distributed variables is again a normally distributed variable, $x_t$ can be sampled directly by a linear combination of $x_0$ and a sample $n'$ from the standard normal distribution, given in equation \ref{newdif}. Having constructed this diffusion process, a machine learning model is trained by getting an input $x_t$ generated using equation \ref{newdif} and predicting $n'$, given $x_t$ and $t$, with the squared error between the model's prediction and $n'$ as loss function. Once this model is trained, data generation is an iterative process. An initial $x_T$ is sampled from the standard normal distribution, the model makes a prediction of $n'$ using this $x_T$ and from this a possible $x_{T-1}$ is found through equations \ref{classicdif} and \ref{newdif}. This process is repeated for the predicted $x_{T-1}$ and then $x_{T-2}$ until finally a synthetic datapoint $x_0$ is generated.
\begin{equation}
\label{newdif}
    x_t = \sqrt{\prod_{i=0}^t[1-\beta_i]}\cdot x_{0} + n' \cdot \sqrt{1-\prod_{i=0}^t[1-\beta_i]}
\end{equation}

\begin{figure}[t]
    \centering
    \includegraphics[width=\linewidth]{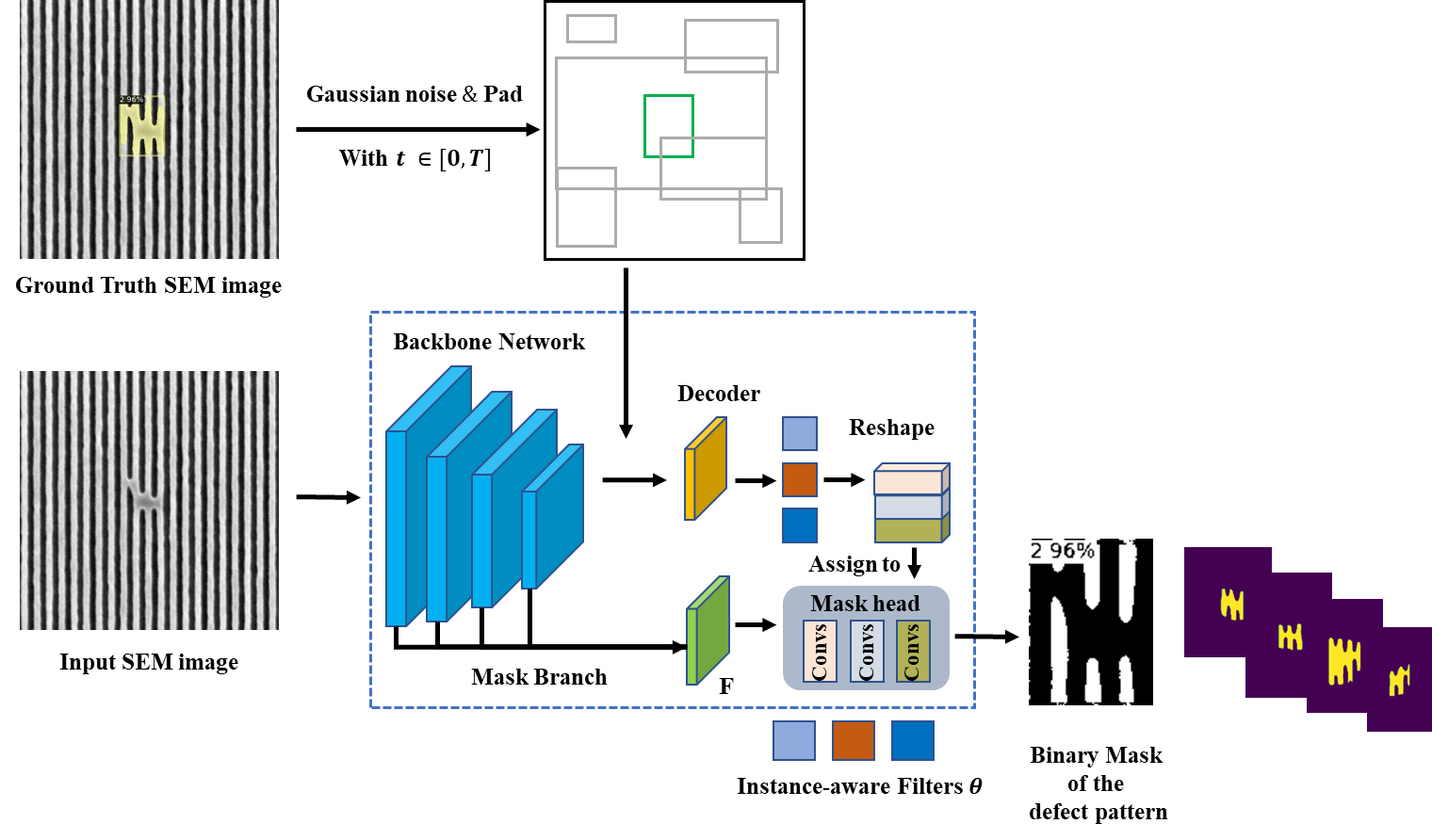}
    \caption{Overview of proposed SEMI-DiffusionInst training framework adapted from \cite{diffinst}.}
    \label{architecture}
\end{figure}

\begin{figure}[b]
    \centering
    \includegraphics[scale=0.4]{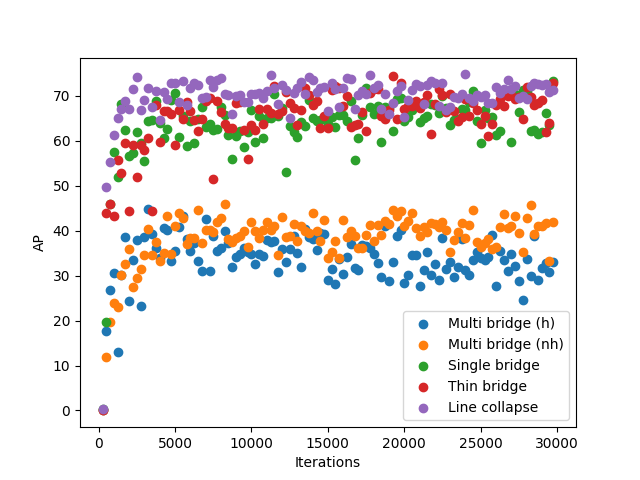}
    \caption{AP metrics for inference on validation dataset between training periods of the ResNet 50 based model.}
    \label{plot}
\end{figure}

\begin{figure*}[t]
    \centering
    \begin{minipage}{\textwidth}
    \includegraphics[width=0.19\linewidth]{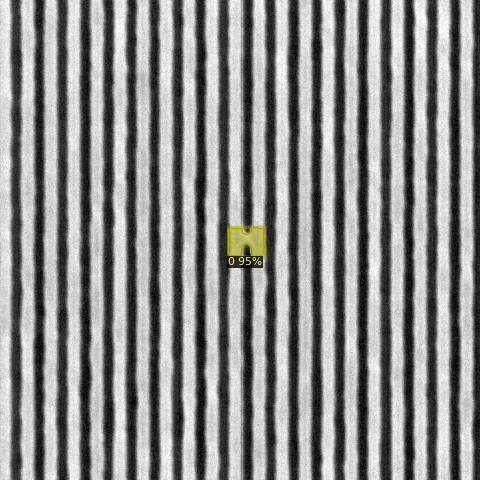}
\hfill
    \includegraphics[width=0.19\linewidth]{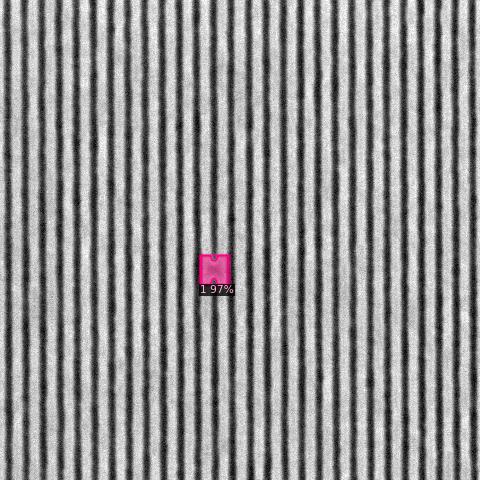}
    \hfill
        \includegraphics[width=0.19\linewidth]{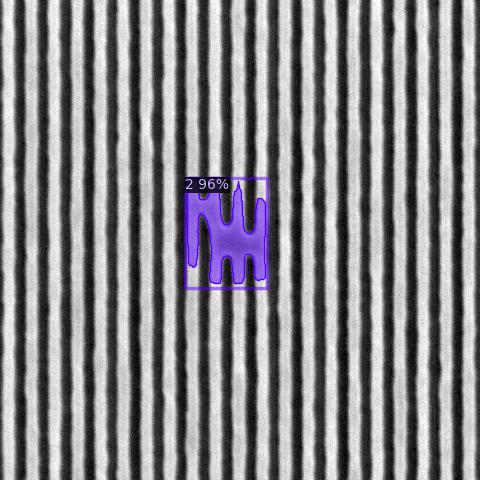}
\hfill
        \includegraphics[width=0.19\linewidth]{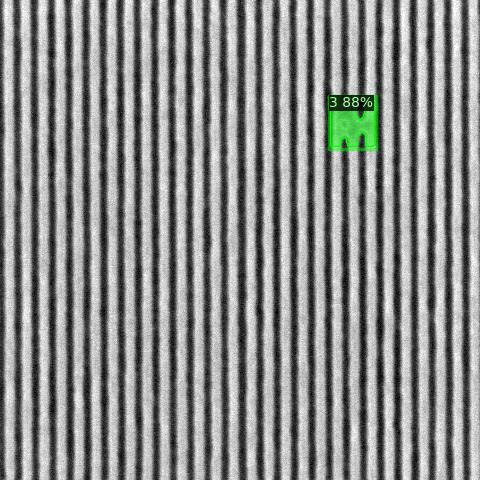}
\hfill
    \includegraphics[width=0.19\linewidth]{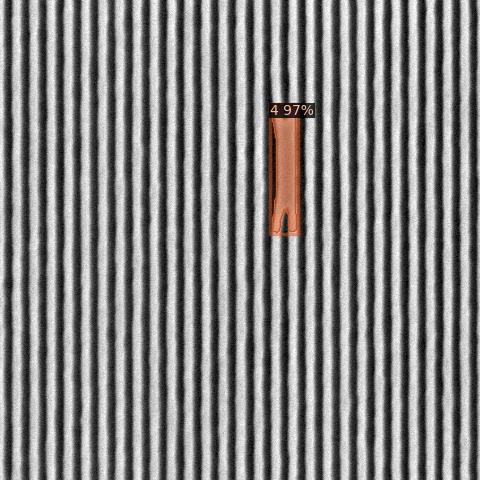}
        \caption{Defect detection and segmentation results with proposed SEMI-DiffusionInst model with Swin backbone and balanced sampling strategy. From left to right: thin bridge [id: 0], single bridge [id: 1], multi bridge (non-horizontal) [id: 2], multi bridge (horizontal) [id: 3], line collapse [id: 4].}
        \label{infresult}
\end{minipage}
\end{figure*}

\subsection{DDPM for object detection and segmentation}\label{expldiffinst}
In \cite{diffdet} the object detection problem is regarded as a generative task by treating it as a denoising diffusion process, training the model to reverse the diffusion process of turning groundtruth bounding boxes into randomly sized and localised boxes. Class probabilities for each box are then predicted using a conventional CNN based technique.

This approach has unique features. To start with, it turns the region proposal and bounding box prediction steps from Mask R-CNN into one task, simplifying the training objective. On top of that, since during inference the number of initial random bounding boxes and inference steps can be freely chosen, it gives the choice of a performance and computational efficiency tradeoff that can be made after the training procedure. 

DiffusionDet has been adapted to the instance segmentation task in DiffusionInst\cite{diffinst}. While the bounding box prediction process remains unchanged, a segmentation mask is predicted along with each bounding box.

In DiffusionInst's predictions, an object mask $m$ is represented as a vector by first obtaining a multiscale fused feature map $F$ using the outputs of different layers of the backbone. Segmentation masks are generated by applying a specific sequence $\phi$ of convolutions to $F$. In \cite{diffinst}, this is a sequence of 1*1 convolutional layers.

\begin{equation}\label{maskeq}
    m = \phi(F,\theta)
\end{equation}

If the mask $m$ is produced using equation \ref{maskeq}, the convolutions' kernel weights $\theta$ are used as a vector representation of $m$. Representing masks this way reduces dimensionality of segmentation masks while ensuring the mask representations are constant throughout different mask map sizes.

\section{Methodology}

\subsection{Dataset}
The dataset used in this work consists of 1040 SEM images of a resist wafer, taken post etch step (After Etch Inspection). The dataset contains various types of stochastic defects, with their distribution shown in Table \ref{tdataset}. Each image is a 480*480 grayscale TIFF image which has been converted to JPG format to comply with the model code. Defects were manually inspected, where segmentation mask, bounding box and class have been annotated for each defect. The dataset was split into 920 images as training dataset and 120 images as validation dataset. During training, images from the training dataset are used with random horizontal flip augmentation to increase the effective dataset size. Use of any synthetic data is avoided to ensure the proposed model's ability to tackle real semiconductor foundry or FAB encountered stochastic defectivity scenarios. These images contain only real stochastic defects (due to process drifts, imaging conditions variations, resist chemical properties etc.) and no intentionally placed or programmed defects. Fig. \ref{exampledataset} shows SEM images with several candidate defect types typically experienced in advanced semiconductor node. These stochastic defect patterns (intra-class level) may appear with variable degrees of pixel-level extent.
\begin{table}[!b]
\renewcommand{\arraystretch}{1.3}
\caption{\textsc{Dataset distribution.}}
\label{tdataset}
\centering
\begin{tabular}{|c||c||c| |c|}
\hline
\textbf{Class name} & \textbf{Id} & \textbf{Training} & \textbf{Validation}\\
\hline
Thin bridge & 0 &  240 & 30 \\
\hline
Single bridge & 1 & 240 & 30 \\
\hline
Multi bridge (Non Horizontal)& 2 & 80 & 10 \\
\hline
Multi bridge (Horizontal)& 3 & 160 & 20 \\
\hline
Line collapse & 4 &200 & 30 \\
\hline
\textbf{Total} & \textbf{/} & \textbf{920} & \textbf{120} \\
\hline
\end{tabular}
\end{table}
\subsection{Backbones}
The application of neural networks, in the paradigm of deep learning, made a revolutionary impact in computer vision domain by outperforming previously practiced manual feature engineering methods (with static kernels/filters). Past few years, Convolution Neural Networks (CNN) and complex architecture variants of it (AlexNet\cite{alexnet}, GoogleNet\cite{goognet}, VGGNet\cite{vgg} etc.) have been extensively used for complex latent feature extraction task, from different domain specific large-scale datasets. These architectures are commonly termed as feature-extractor networks or backbones and are capable of extracting many hidden low-level, mid-level and high-level features from images using dynamic kernels.
Fig. \ref{architecture} illustrates the DiffusionInst model based automated defect classification, detection and segmentation (ADCDS) framework, as SEMI-DiffusionInst, which is trained and evaluated using selected backbones on imec dataset (SEM images with candidate defect types as discussed in previous section). Our implementation is adapted from the publicly-available GitHub repository \cite{diffinst}.

The backbones  used in this research work are convolution based ResNet 50\cite{resnet}, ResNet 101\cite{resnet} and transformer based Swin\cite{swin} which were all pretrained on the Imagenet dataset\cite{imagenet}. Performance metrics of these 3 feature extractor networks/backbones are investigated and compared. ResNet 50 and ResNet 101 based models are trained with batchsize of 8, while Swin is trained with batchsize of 2 due to gpu memory limits. For fair comparison ResNet based models went through 30000 training iterations while the Swin based model went through 120000 iterations. Learning rate and optimizer settings are the same compared to the ones used for COCO dataset in \cite{diffinst}.

\subsection{Training data sampling strategy}
Table \ref{tdataset} shows a class imbalance scenario among the different defect types. Multi-bridge defects are underrepresented in the dataset. Therefore the performance difference is investigated between different backbones (as ResNet 50, ResNet 101 and Swin) trained using a uniform data sampling strategy and backbones trained with a weighted data sampling strategy. Following this strategy, weighting occurs to guarantee that all defect types must be encountered in approximately 20\% of training images. Since the dataset has 5 unique defect classes and each image contains only 1 defect pattern, the maximum frequency $f$ with which each defect type may appear is $f \approx \frac{100}{n=5}$\%, where n is the number of defect classes. 

\subsection{Inference Random Boxes}
As explained in section \ref{expldiffinst}, once a model has been trained it can be used during inference with varying number of initial random boxes. Increasing this number signifies extensive opportunities to find the exact defect region, against additional computation. While \cite{diffdet} has suggested 500 random initial boxes as the optimal tradeoff between speed and performance, our dataset contains fewer classes and has less diverse backgrounds. Thus it is possible to reach an optimal performance with a lower number of initial random boxes. This is investigated alongside the speed (inference time) gained from lowering the random box number.

\begin{table*}[b]

\renewcommand{\arraystretch}{1.3}
\caption{\textsc{Best per category bounding box APs achieved on validation/test dataset, averaged over predictions
using IOU thresholds between 0.5 and 0.95 with 0.05 step size. Best, second best and third best colour codes: \textcolor{green}{Green}, \textcolor{red}{Red}, \textcolor{blue}{Blue}}.}
\label{t_bestb}
\centering
\begin{tabular}{|c||c||c||c||c||c||c||c|}
\hline
\textbf{Backbone} & \textbf{Sampling strategy} & \textbf{Line Collapse} & \textbf{Single Bridge} & \textbf{Multi bridge (NH)} & \textbf{Thin bridge} & \textbf{Multi bridge (H)}& \textbf{mAP}\\
\hline
ResNet 50 & Standard & 81.06 & 74.75 & 56.12 & 78.77 & 49.83 & 64.24\\
\hline
ResNet 101 & Standard & 79.89 & \textcolor{blue}{75.56} & 58.89 & \textcolor{blue}{81.23} & \textcolor{blue}{53.80} & \textcolor{blue}{66.83}\\
\hline
Swin & Standard & \textcolor{red}{82.44} & 75.04 & \textcolor{red}{62.36} & \textcolor{green}{\textbf{83.08}} & \textcolor{green}{\textbf{57.23}} & \textcolor{green}{\textbf{67.82}} \\
\hline
ResNet 50 & Balanced & 79.83 & 72.30 & 57.87 & 79.33 & 50.63 & 64.11 \\
\hline
ResNet 101 & Balanced & \textcolor{blue}{81.71} & \textcolor{green}{\textbf{78.32}} & \textcolor{blue}{60.63} & 81.10 & \textcolor{red}{55.18} & 66.00 \\
\hline
Swin & Balanced & \textcolor{green}{\textbf{83.23}} & \textcolor{red}{76.97} & \textcolor{green}{\textbf{63.25}} & \textcolor{red}{81.25} & 53.39 & \textcolor{red}{67.39} \\
\hline
\end{tabular}
\end{table*}
\subsection{Metrics}
To compare different backbones of the proposed SEMI-DiffusionInst model and other metrics, inference time is measured using the Detectron2 library\cite{detectron2} by averaging the pure compute time over all batches used during inference on the entire validation dataset with batchsize 1.

To evaluate the defect classification and segmentation performance, per defect class AP (Average Precision) and overall mAP (mean Average Precision) have been used, both being an average of the AP achieved using predictions with an Intersection Over Union (IOU) range of 0.5 to 0.95, an IOU step size of 0.05 and a confidence threshold score of 0.5. Proposed SEMI-DiffusionInst model with (a) 3 different backbones and (b) sampling strategy is periodically evaluated on the validation dataset, and the best APs (mAP on entire validation dataset and per defect class AP) have been recorded. Fig. \ref{plot} demonstrates AP metrics for inference on validation dataset between training periods of the ResNet 50 based model.

\section{Results and discussion}

\subsection{Backbones}

\begin{table*}[!t]

\renewcommand{\arraystretch}{1.3}
\caption{\textsc{Best per category segmentation APs achieved on validation/test dataset, averaged over predictions
using IOU thresholds between 0.5 and 0.95 with 0.05 step size. Best, second best and third best colour codes: \textcolor{green}{Green}, \textcolor{red}{Red}, \textcolor{blue}{Blue}.}}
\label{t_bests}
\centering
\begin{tabular}{|c||c||c||c||c||c||c||c|}
\hline
\textbf{Backbone} & \textbf{Sampling strategy} &\textbf{Line Collapse} & \textbf{Single Bridge} & \textbf{Multi bridge (NH)} & \textbf{Thin bridge} & \textbf{Multi bridge (H)}& \textbf{mAP}\\
\hline
ResNet 50 & Standard & \textcolor{blue}{74.76} & 73.33 & 46.03 & \textcolor{blue}{74.33} & 44.80 & 59.76\\
\hline
ResNet 101 & Standard & 73.68 & \textcolor{red}{74.31} & \textcolor{blue}{49.03} & 74.09 & \textcolor{blue}{47.64} & 61.05\\
\hline
Swin & Standard & 74.28 & 74.22 & \textcolor{green}{\textbf{50.75}} & \textcolor{green}{\textbf{75.38}} & \textcolor{green}{\textbf{53.09}} & \textcolor{green}{\textbf{63.01}}\\
\hline
ResNet 50 & Balanced & 73.30 & 71.84 & 46.04 & 72.56 & 44.92 & 58.99\\
\hline
ResNet 101 & Balanced & \textcolor{red}{75.37} & \textcolor{green}{\textbf{77.60}} & 48.72 & 73.79 & \textcolor{red}{51.84} & \textcolor{blue}{61.80}\\
\hline
Swin & Balanced &\textcolor{green}{\textbf{75.50}} & \textcolor{blue}{74.29} & \textcolor{red}{50.21} & \textcolor{red}{75.07} & 47.59 & \textcolor{red}{61.86}\\
\hline
\end{tabular}
\end{table*}

\begin{table*}[t]

\renewcommand{\arraystretch}{1.3}
\caption{\textsc{Comparison analysis of defect segmentation and detection using proposed SEMI-DiffusionInst vs SEMI-PointRend\cite{pointrend}.}}
\label{compare}
\centering
\begin{tabular}{|c||c||c||c||c||c||c|}
\hline
\textbf{Model\&Task} & \textbf{Line Collapse} & \textbf{Single Bridge} & \textbf{Multi n-h bridge} & \textbf{Thin bridge} & \textbf{Multi h bridge}& \textbf{mAP}\\
\hline
SEMI-PointRend BBox\cite{pointrend}& 68.7  & \textbf{76.5} & \textbf{69.6} & 58.8 & 52.8 & 65.3\\
\hline
\textbf{Proposed SEMI-DiffusionInst BBox} & \textbf{82.44} & 75.04 & 62.36 & \textbf{83.08} & \textbf{57.23} & \textbf{67.8}\\
\hline
SEMI-PointRend segmentation\cite{pointrend} & 63.6 & \textbf{77.7} & \textbf{57.4} & 55.0 & \textbf{53.5} & 61.7\\
\hline
\textbf{Proposed SEMI-DiffusionInst segmentation} & \textbf{74.28} & 74.22 & 50.75 & \textbf{75.38} & 53.09 & \textbf{63.0}\\
\hline
\end{tabular}
\end{table*}
Tables \ref{t_bestb} and \ref{t_bests} show the best APs achieved during validation/test by proposed SEMI-DiffusionInst model with different backbones on defect classification, detection and segmentation task. Fig. \ref{infresult} shows defect detection and segmentation results with proposed SEMI-DiffusionInst model with Swin backbone and balanced sampling strategy. While proposed SEMI-DiffusionInst model with a Swin backbone significantly outperforms both ResNet 50 and ResNet 101 based models, the Swin backbone based model requires more computational resources and parameters. Table \ref{modelstats} shows the memory required to store model weights and corresponding model inference time per image using one sampling step and 500 random boxes on a NVIDIA Tesla V100 GPU.

Comparing backbones, a substantial performance increase can be observed from ResNet 50 based models to ResNet 101
or Swin based models, while the latter two backbone based models perform as per. However, as shown in Table \ref{modelstats}, ResNet 101 based models are significantly faster and smaller than Swin based models, which suggests ResNet 101 as the ideal backbone for our dataset for a speed-performance tradeoff.

Another perspective can be drawn as, regardless of different backbones, all models could not perform well on the multi bridge defect classes (both horizontal and non-horizontal) compared to other defect classes. One possible reason can be the complex pattern shape and larger intra-class variation for these particular defect class to uniquely learn from limitedly available data. Future work can be extended towards (a) training from scratch or (b) using backbones pretrained on SEM images which may lead to better latent feature extraction during training rather than using ImageNet pretrained backbones and fine-tune them. Therefore, a more optimal value for "loss-function" can be found during training with the same datapoints, which may potentially lead to substantial performance gain on multi bridge defect class (as well as other defect patterns) combined with different other strategies as discussed in next sections.
\begin{table}[!b]
\renewcommand{\arraystretch}{1.3}
\caption{\textsc{Model metrics.}}
\label{modelstats}
\centering
\begin{tabular}{|c||c||c|}
\hline
\textbf{Backbone} & \textbf{Parameters} & \textbf{Inference (s/image)} \\
\hline
ResNet 50 & 1.315GB & 0.0964 \\
\hline
ResNet 101 & 1.537GB & 0.1084 \\
\hline
Swin & 2.053GB & 0.1439 \\
\hline
\end{tabular}
\end{table}

\setlength{\textfloatsep}{1mm}
\begin{table}[!b]
\renewcommand{\arraystretch}{1.3}
\caption{\textsc{Results for different random box numbers running a ResNet 50 based model on validation dataset.}}
\label{randbox}
\centering
\begin{tabular}{|c||c||c|}
\hline
\textbf{Random boxes} & \textbf{mAP} & \textbf{Inference (s/image)}\\
\hline
500 & 62.68 & 0.1084 \\
\hline
400 & 63.07 & 0.0889 \\
\hline
300 & 62.81 & 0.0786 \\
\hline
200 & 62.79 & 0.0695 \\
\hline
100 & 62.13 & 0.0573 \\
\hline
50 & 59.89 & 0.0518 \\
\hline
25 & 53.81 & 0.0494 \\
\hline
\end{tabular}
\end{table}
\setlength{\textfloatsep}{2mm}
\subsection{Sampling strategies}
While the balanced sampling strategy has a negligible impact on overall mAP metric, ResNet 101 and Swin based models do benefit slightly on certain defect types. For segmentation AP, the ResNet 101 based model trained with balanced sampling outperforms conventionally trained ResNet 101 (with uniform sampling) based model for line-collapse (2.29\%), single bridge (4.43\%) and multi-bridge horizontal defects (8.82\%), respectively. For the same, Swin based model with balanced sampling achieves 1.64\% improvement for line-collapse and performs as per (second best or third best) for almost all other defect types. The same trend can be followed for bbox AP for these two variants. Our goal of this research work is to experiment with different strategies (such as model centric approach with different backbones, sampling rate, random box initialization etc.) to tackle the challenge of limited data availability and/or class imbalance scenario without using any biased/synthetic data. The main problem with synthetic dataset/images, generated from arbitrary software/tools, is, those do not satisfy the requirements of the semiconductor industry at advanced nodes and often may lead to additive artefacts. The finding of our experiment indicates that number of multi bridge defect instances (both horizontal and non-horizontal) are not sufficient to achieve similar precision on these defects compared to other defect types. In an actual semiconductor foundry, it is reasonably obvious that some defect class (for an example here, multi bridge) is not encountered at the same frequency as other defect classes, since these defects are stochastic. Therefore, future work can be extended towards combining this modified (balanced) sampling method with advanced data augmentation techniques to intrinsically increase the used dataset.

\subsection{Random Boxes}
Table \ref{randbox} demonstrates our experimentation with number of random boxes initialization during inference time. While 500 initial random boxes is used in \cite{diffdet, diffinst} as a balanced tradeoff between speed and precision to investigate their models on the COCO and LVIS datasets, this does not guarantee optimal performance on our SEM based defect dataset. As a proof-of-concept, we have chosen a range of the number of random boxes from 500 to 25. Table \ref{randbox} shows that decreasing the number of random boxes from 500 to 100 does not deteriorate performance (mAP) much, rather indicative towards an improved performance compared to 500 random boxes up to 200 random boxes. Also, it is evident that decreasing to 100 random boxes causes a significant improvement (47.14\%) in inference time. This experimentation leads to the fact that, since \cite{diffdet, diffinst} investigated their models on the COCO and LVIS datasets, their recommendations for optimal hyperparameters during both training and inference will not transfer perfectly
to the studied semiconductor dataset.

\subsection{Comparison to previous work}
Table \ref{compare} compares the performance on both the defect pattern segmentation and detection task on the same dataset of the SEMI-PointRend model \cite{pointrend} to the proposed SEMI-DiffusionInst model with Swin backbone as feature extractor network. While the Swin based SEMI-DiffusionInst model achieves better overall mAP on both tasks, it somewhat underperforms on multi bridge (non-horizontal) defect pattern (11.58\% drop in segmentation precision compared to the SEMI-PointRend model). However, the proposed approach outperforms (or as per) the SEMI-PointRend model for thin bridge (37.5\%), line collapse (16.79\%), single bridge and multi bridge (horizontal) defect patterns. While the Swin based SEMI-DiffusionInst model is less effective at single bridges compared to SEMI-PointRend, ResNet 101 (with balanced sampling) based SEMI-DiffusionInst model performs as per to SEMI-PointRend on single bridge defects. These findings open up another research direction as future experimentation strategy to implement an ensemble model based on a selective permutation of backbones (combining predictions from ResNet 50, ResNet 101, Swin, vgg-19 etc.).

\section{Conclusion}
In this research work, the DiffusionInst model is investigated for semiconductor defect inspection, specifically for precise pixel-wise segmentation of defect patterns. While benchmarking DiffusionInst models with different backbones, ResNet-101 based DiffusionInst models offered the best trade-off in terms of precision and computational efficiency. Models were trained first with a standard data sampler, but to tackle the class imbalance scenario in the dataset, models were trained another time with a modified sampler which takes the defect type frequencies into account.

Finally, results between the proposed SEMI-DiffusionInst model and previous SEMI-PointRend model are compared, where our proposed approach is better or as per for almost all defect classes (per class APs) as well as outperforms mAPs. The bounding box and segmentation mAPs achieved by the SEMI-DiffusionInst model are improved by 3.83\% and 2.10\%, respectively. Among individual defect types, precision on line collapse and thin bridge defects are improved the most, with approximately 15\% on detection task for both defect types.

Future work can be extended to the finetuning of learning rate and finding other optimal hyperparameters, since this work used learning rates optimized for the COCO dataset, which may not be optimal for the investigated semiconductor wafer dataset. Furthermore, architectural changes to the SEMI-DiffusionInst segmentation mask head can be investigated to enhance the predicted segmentation mask precision. Finally, experimenting with datasets containing more multi bridge horizontal and non-horizontal defects may be proven to be rewarding to increase precision on these defects, since all investigated backbones demonstrate a signficantly lower performance on these defects compared to the others, which may partly be caused as lower number of these defect instances were present in the training data.

\bibliographystyle{ieeetr}
\bibliography{reference}

\end{document}